\pdfoutput=1

\documentclass[11pt]{article}

\usepackage{EACL2023}

\usepackage{times}
\usepackage{latexsym}

\usepackage[T1]{fontenc}

\usepackage[utf8]{inputenc}

\usepackage{microtype}

\usepackage{inconsolata}

\usepackage{enumitem} 
\usepackage{xcolor}
\usepackage{colortbl}

\usepackage{soul}

\newcommand{\sm}[1]{{\color{black} #1}} 

\title{Best Practices in the Creation and Use of Emotion Lexicons}

\author{Saif M. Mohammad\\
	    National Research Council Canada\\
	     {\tt saif.mohammad@nrc-cnrc.gc.ca} }

\begin{document}
\maketitle
\begin{abstract}
Words play a central role in how we express ourselves. Lexicons of word--emotion associations are widely used in research and real-world applications for sentiment analysis, tracking emotions associated with products and policies, studying health disorders, tracking emotional arcs of stories, and so on. However, inappropriate and incorrect use of these lexicons can lead to not just sub-optimal results, but also inferences that are directly harmful to people. This paper brings together ideas from Affective Computing and AI Ethics to present, some of the practical and ethical  considerations involved in the creation and use of emotion lexicons ---  \textit{best practices}. The goal is to provide a comprehensive set of relevant considerations, so that readers (especially those new to work with emotions) can find relevant information in one place.
We hope this work will facilitate more thoughtfulness when one is deciding on what emotions to work on, how to create an emotion lexicon, 
how to use an emotion lexicon, how to draw meaningful inferences, and how to judge success. 
\end{abstract}


\section{Introduction}
\label{sec:intro}
 \setitemize[0]{leftmargin=*}
 \setenumerate[0]{leftmargin=*}


\noindent Words often convey affect (emotions, sentiment, feelings, and attitudes); either explicitly through their core meaning (denotation) or implicitly through connotation. 
For example, \textit{dejected} denotes sadness. 
On the other hand, \textit{failure} simply connotes sadness. 
Either through denotation or connotation, both words are associated with sadness. 
A compilation of such associations is referred to as a {\it word--affect association lexicon} (aka \textit{emotion lexicon}).\footnote{This includes
\textit{sentiment lexicons} that capture valence (association with the positive--negative dimension) and other lexica that capture affect-related
phenomena.}
An entry in a lexicon usually includes a word, an emotion category or affect dimension (e.g., joy, fear, valence, arousal, etc.), and a score indicating association (or strength of association). 

Examples of emotion lexicons include the General Inquirer \cite{Stone66}, ANEW \cite{nielsen2011new,bradley1999affective}, LIWC \cite{pennebaker2001linguistic},
Pittsburgh Subjectivity Lexicon \cite{wilson2005recognizing}, \textit{NRC Emotion Lexicon} \cite{MohammadT10,MohammadT13}, 
and the \textit{NRC Valence, Arousal, and Dominance (VAD) Lexicon} \cite{vad-acl2018}.
These were all created by manual annotation (either by experts or crowdsourced).
There also exist lexicons that were generated automatically from large text corpora using  statistical and/or machine learning algorithms; e.g., WordNet Affect \cite{strapparava2004wordnet}, SentiWordNet (SWN) \cite{baccianella2010sentiwordnet}.

Emotion lexicons have 
a wide range of applications in commerce, public health, and research (in NLP, Psychology, Social Sciences, Digital Humanities, etc.). 
Some notable examples include: tracking brand and product perception via social media posts, tracking support for controversial issues and policies, tracking buy-in for non-pharmaceutical health measures such as social distancing during a pandemic, literary analysis, and developing more natural dialogue systems. 
The lexicons can be used on their own or in support of neural machine learning (ML) algorithms for emotion recognition.


Lexicon-based emotion analyses are especially popular in real-world applications and research outside of computer science because they are interpretable, have a low carbon footprint, and do not require significant programming
expertise. Further, since outputs of ML models are highly dependent on training data, use of a model often requires retraining, and there may not exists labeled data from the target domain
Further, \newcite{TMemotionarcs} show that when determining broad trends (emotion arcs) and aggregating information from hundreds (if not more) instances for every time step, simple lexicon-based methods are extremely accurate (correlations above 0.95 with ground truth arcs).

However, inappropriate and incorrect use of these lexicons, can lead to not just sub-optimal results, but also inferences that are directly harmful. \sm{For example, using lexicons to infer emotions from limited amount of data to make judgments about refugee applications, 
to make judgments about which groups of people are shown certain advertisements and which groups are not, 
marking businesses owned by some groups of people as less liked than that of others, etc.}

Emotions are deeply personal, private, and complex. Even the best natural language systems largely only employ pattern matching based on huge amounts of historical data, and thus often do not really understand what the user is 
trying to convey, let alone how they are 
feeling. 
In fact, some 
recent commercial and governmental uses of emotion recognition have garnered considerable criticism, including:
 infringing on one's privacy, exploiting vulnerable sub-populations, and 
 even allegations of pseudo-science \cite{Mohammad22AER,wakefield_2021,article19_2021,woensel_nevil_2019}.

This paper brings together ideas from Affective Computing and AI Ethics to present, in one place, some of the practical and ethical  considerations involved in the creation and use of emotion lexicons ---  \textit{best practices}.\footnote{This paper is a reframed and expanded avatar of an earlier datasheet paper for emotion lexicons  \cite{mohammad2020practical}.} 
We hope this work will facilitate more thoughtfulness when one is deciding on what emotions to work on, how to create an emotion lexicon, 
how to use an emotion lexicon, and how to judge success. 
Additional benefits of such a document include:
\vspace*{-1mm}
\begin{enumerate}
    \item Presents the trade-offs of relevant choices so that stakeholders can make informed decisions appropriate for their context. \\[-16pt]
    \item Has citations and pointers; acts as a jumping off point for further reading.\\[-16pt]
    \item Helps engage the various stakeholders of an emotion task with each other. Helps stakeholders challenge assumptions made by researchers and developers.\\[-16pt]  
     \item Helps develop harm mitigation strategies.\\[-16pt]
    \item Acts as a useful introductory document on emotion lexicons (complements survey articles).
\end{enumerate}
\vspace*{-1mm}
\noindent Note that even though this article is focused on emotion lexicons, many of the ethical considerations apply broadly to natural language lexicons/resources in general. 
Also, see \newcite{Mohammad22AER} for a broader discussion on the ethical considerations associated with automatic emotion recognition (AER).

\sm{This work is in the same spirit as other recent innovations in exercising responsible research such as
datasheets for datasets \cite{Gebru2018DatasheetsFD}, model cards for systems \cite{mitchell2019model},
and ethics sheets for AI tasks \cite{mohammad-2022-ethics}. However, unlike datasheets and model cards which are designed for individual 
datasets and systems and that are published after the work is done, the goal of this work is to provide a more general-purpose relevant resource,
accessible at the very beginning of one's project.
Also, unlike an ethics sheet for a automatic emotion recognition that may cover all kinds of ethical considerations associated with the task of interest, this document has a focus on  the creation of emotion lexicons and their use in AI tasks. 
}

\sm{Ethics considerations are not about objective metrics or simple checklists. They involve engaging with issues that impact stake holders, especially those that are already disadvantaged. Thus, a big component of this work is to raise awareness of relevant issues, to underscore how often there are no easy solutions, and that meaningful change requires painstaking, slow, and deliberate engagement with the stakeholders. 
Additionally, 
such documents are useful for those that are impacted to question and challenge assumptions made by unfair decisions of automated systems.}

\section{Best Practices}
\noindent Below we present various best practices (practical and ethical considerations) pertaining to 22 aspects of emotion lexicon creation and use. The 22 aspects are grouped under the coarser categories 
pertaining to a lexicon's life cycle:
A. Lexicon Design, B. Annotation, C. Entries in the Lexicon, and D. Applying the Lexicon. 
Note that while many considerations are presented from the perspective of lexicon creation, they are also relevant to the users of a lexicon --- knowing what decisions were made during the creation of a lexicon help one to assess appropriateness of using the lexicon. 

\sm{The goal 
is to provide a comprehensive set of relevant considerations, so that readers (especially those new to research or new to work with emotions) can find 
the information in one place. Thus, we include both the considerations that are especially specific to emotions, as well as others that apply more broadly (even if they are somewhat well known).} 
Also, the points listed below are not meant to be  
the final word,
but rather jumping off points for further thought and discussion. 

\subsection{Overview}

\noindent An overview of the 22 aspects is presented below; followed by the detailed descriptions.\\[-10pt] 

\noindent A. LEXICON DESIGN\\[2pt]
\hspace*{3mm}  1. Purpose or Objective\\
\hspace*{3mm}  2. Emotion Category or Dimension\\
\hspace*{3mm}  3. Word Senses and Dominant Sense Priors\\
\hspace*{3mm}  4. Discrete or Continuous Value Labels\\[4pt]
\noindent B. ANNOTATION\\[2pt]
\hspace*{3mm}  5. Questionnaire\\[2pt]
\hspace*{3mm}  6. Comparative Annotations\\ 
\hspace*{3mm}  7. Annotators\\
\hspace*{3mm}  8. Quality Control \\ 
\noindent C. ENTRIES IN THE LEXICON\\[2pt]
\hspace*{3mm}  9. Annotation Aggregation\\
\hspace*{3mm}  10. Relative (not Absolute)\\
\hspace*{3mm}  11. Coverage\\
\hspace*{3mm}  12. Not Immutable\\
\hspace*{3mm}  13. Perceptions (not ``truth'')\\
\hspace*{3mm}  14. Socio-Cultural Biases\\
\hspace*{3mm}  15. Inappropriate Biases\\
\hspace*{3mm}  16. Errors\\
\hspace*{3mm}  16. Mechanism to Report and Fix Errors\\[4pt]
\noindent D. APPLYING THE LEXICON\\[2pt]
\hspace*{3mm}  18. Fit of the Lexicon to One's Data\\
\hspace*{3mm}  19. Rescaling the Lexicon for One's Task\\
\hspace*{3mm}  20. Metrics \& Features Drawn from the Lexicon\\
\hspace*{3mm}  21. Removing Neutral Words\\
\hspace*{3mm}  22. Inferences \\

\vspace*{-5mm}
\subsection{Detailed Descriptions}

\noindent \textbf{A. LEXICON DESIGN}\\[-8pt]

\noindent \textbf{\#1. Purpose or Objective:} Consider and document the objective(s) of building the emotion lexicon. There can be more than one objective. 
The objectives guide various design choices involved in the creation of the lexicon. 
 See \newcite{selbst2019fairness} for common pitfalls in designing and framing socio-technical systems; and \newcite{Mohammad22AER} for common pitfalls in designing and framing automatic emotion recognition tasks.  Users of emotion lexicons can study the purpose of each lexicon to determine which is most suitable for their use case.

Broadly speaking, the objectives tend to be around the study of word--emotion associations (exploring various research questions at the intersection of language an emotions) and aiding automatic emotion detection from utterances. However, 
individual
projects often have specific goals, for example, 
to study specific phenomenon such as loneliness and empathy, to study inappropriate biases, to detect what emotions people perceive from utterances, 
to study how automatic systems should perceive the emotions in utterances, 
how automatic systems should use words to convey emotions, etc. It is important to recognize that some of these objectives are very related, but they have important differences. For example, while a general-purpose emotion lexicon will capture a number of benign associations, it will also capture inappropriate societal biases. If one wants to use a lexicon in a text generation system, then they should either use a lexicon designed specifically for that purpose, or address the biases in a general purpose lexicon, before using it.

Work using emotion lexicons should not claim that using it one can determine one's emotional state from their utterance. At best, recognition systems (whether they use emotion lexicons or not) capture what one is trying to convey or what is perceived by the listener/viewer; and even there, given the complexity of human expression, they are often inaccurate.
Several studies have shown that it is difficult to fully measure psychological states of people \cite{stark2018algorithmic,barrett2017theory}.

In contrast, statistical analyses with features drawn from emotion lexicons can be used to accurately determine broad trends in the emotional state of a population over time \cite{TMemotionarcs}. Here, inferences are drawn at aggregate level from much larger amounts of data. 
Studies on public health, such as those on loneliness \cite{guntuku2019studying,kiritchenko-etal-2020-solo}, depression \cite{de2013predicting,resnik-etal-2015-beyond}, suicidality prediction \cite{macavaney-etal-2021-community}, bipolar disorder \cite{karam2014ecologically}, stress \cite{eichstaedt2015psychological}, 
emotions during a pandemic \cite{VM2022-TED},
and general well-being \cite{rtz2013cschwaharacterizing} fall in this category. Here too, however, it is best to be cautious in making claims about mental state, and use emotion recognition as one source of evidence amongst many (and involve expertise from public health and psychology). 

\noindent {\bf \#2. Emotion Category or Dimension:} A key decision in the creation of an emotion lexicon is 
which conceptualization or facet of emotion to use.
For example, should it capture emotion categories such as joy, sadness, fear, optimism, etc., or will it capture dimensions such as valence, arousal, and dominance.
Psychologists and neuro-scientists have identified several theories of emotion that can inform the choice of categories and dimensions, including: the Basic Emotions Theory (BET) \cite{ekman1992there,ekman1994nature}, the Dimensional Theory \cite{osgood1957measurement,russell1980circumplex,russell1977evidence,russell2003core},
Cognitive Appraisal Theory \cite{scherer1999appraisal,lazarus1991progress}, and the Theory of Constructed Emotions \cite{barrett2017theory}.

Since ML approaches rely on human-annotated data (which can be hard to obtain in large quantities), emotion recognition research has often gravitated to the Basic Emotions Theory, as that work allows one to focus on a small number of emotions. This attraction has been even stronger in the vision research because of BET's suggested mapping between facial expressions and emotions. However, many of the tenets of BET, such as the universality of some emotions and their fixed mapping to facial expressions, stand discredited or are in question \cite{barrett2017emotions,barrett2019emotional}.


Carefully consider which emotion formulation you wish to capture in your lexicon, or is appropriate for your task/project. For example, one may choose to work with the dimensional model or the model of constructed emotions if the goal is to infer behavioural or health outcome predictions. 
Despite criticisms of BET, it makes sense for some NLP work to focus on \textit{categorical emotions} such as joy, sadness, guilt, pride, fear, etc.\@ (including what some refer to as basic emotions) because people often talk about their emotions in terms of these concepts. Many human languages have words for these concepts (even if our individual mental representations for these concepts vary to some extent) \cite{wierzbicka1999emotions}. However, note that work on categorical emotions by itself is not an endorsement of the BET. Do not refer to some emotions as basic emotions, unless you mean to convey your belief in the BET. Careless endorsement of theories can lead to the perpetuation of ideas that are actively harmful (such as suggesting we can determine internal state from outward appearance---physiognomy).\\[-8pt]


\noindent \textbf{\#3. Word Senses and Dominant Sense Priors:} Words when used in different senses and contexts may be associated with different emotions. The entries in the emotion lexicons are mostly indicative of the emotions associated with the predominant senses of the words. This is usually not too problematic because most words have a highly dominant main sense (which occurs much more frequently than the other senses). 
In specialized domains, some terms might have a different dominant sense than in general usage. Entries in the lexicon for such terms should be appropriately updated or removed. However, if the goal of the project is to create a lexicon for a specialized domain, then one should guide the annotation process accordingly.\\[-8pt]

\noindent {\bf \#4. Discrete or Continuous Value Labels:} Many emotion lexicons have discrete binary labels for words (positive--negative, joy--no joy, fear--no fear, and so on). Lexicons such as ANEW and the NRC VAD Lexicon 
have real-valued scores between 0 and 1, -1 and 1, 0 to 5, 0 to 100, etc.
Real-valued scores allows one to make finer distinctions in the degree of emotion. They allow one to determine the intensity of emotion. Binary-labeled lexicons are used primarily to determine density of emotion word usage; for example,
to explore whether there is a higher percentage of tweets with loneliness words during the Covid-19 pandemic, than in the years before the pandemic. 
Determine 
which type of lexicon is more aligned with your objectives. \\[4pt] 

  
 
 
 
 

\noindent \textbf{B. ANNOTATION}\\[-8pt]

\noindent {\bf \#5. Questionnaire:} Arguably the most crucial aspect in the creation of an emotion lexicon is the questionnaire. What is asked and how it is asked determines the outcome. Below are key recommendations in the design of questionnaires:\\[-16pt]
\begin{enumerate}[label=\alph*.]
\item Where appropriate, break the task/question into simpler sub-tasks/sub-questions.\\[-18pt]
\item It is better to have separate tasks for different questions and emotion dimensions. Asking for responses about more than one emotion dimension requires the annotator to switch contexts and leads to more cognitive load.\\[-18pt]
\item Keep the instructions clear and easy to follow.\\[-18pt]
\item Examples are more important than definitions. People tend to learn faster and better through examples. It is still good to include simple definitions of relevant concepts.\\[-18pt]
\item Refer to the theories for emotions work in psychology on to how to collect emotional information from respondents. 
Especially useful are the terms used 
to define emotion dimensions:
e.g., as per the dimensional model of emotions \cite{russell1980circumplex} \textit{arousal} is defined as the active--sluggish dimension,
in the stereotype content model of social perception \cite{cuddy2008warmth}, \textit{warmth} is defined as the trustworthiness, friendliness, kindness dimension. These words should be used when eliciting annotation responses.\\[-18pt] 
\item Keep the instructions brief. This is respectful of annotator time, and one can only keep track of a limited number of instructions at a time.\\[-18pt]
\item Explain the purpose of the annotation task. This is respectful of annotators. 
 People have a right to know (in appropriate detail) what research they are contributing their time for. This may also lead to more engaged annotators.  \\[-18pt]
\item Include an optional comment box that gives annotators a way to provide feedback, raise issues, and to be heard.\\[-18pt]
\item Make the questionnaire and instructions freely available. 
This helps others 
to build on your work. It allows users to see exactly how the questions were phrased, and thus how to interpret the resulting emotion lexicon.\\[-18pt]
\end{enumerate}
\noindent See also other data curation and questionnaire development tips from non-NLP fields such as psychology \cite{doi:10.1177/1094428119836485}. \\[-4pt]


\noindent {\bf \#6. Comparative Annotations:} Real-valued scores provide fine-grained emotion information; 
however, it is difficult for humans to provide direct scores at this granularity. 
A popular approach to obtain real-valued scores is by providing the annotators with numeric rating scales.\footnote{https://www.questionpro.com/blog/rating-scale/}
These scales have numbers (usually 1 to 5 or 1 to 7) and the annotator has to select which number is most indicative of the degree of association with the property of interest for the given word; given that the lowest number on the scale indicates least association and the highest number indicates the most association.\footnote{It is good practice to anchor the numeric values with 
labels such as maximum/moderate/low association.}
The scores for an item from multiple annotators is averaged to obtain a real-valued score that is assigned to the word--emotion pair. 

A common problem of annotation by rating scales is inconsistencies in annotations among different annotators. One annotator might assign a score of 87 to one word, while another annotator may assign a score of 81 to the same word. It is also common that the same annotator might assign different scores to the same word, 
if asked to annotate again after a period of time.
Further, annotators often have a bias towards selecting scores in the middle of the scale, known as {\it scale region bias} \cite{presser2004questions,baumgartner2001response}.\\[-10pt]

\noindent {\it Paired Comparisons} \cite{thurstone1927law,david1963method} is a comparative annotation method, 
where respondents are presented with pairs of items and asked which item has more of the property of interest (for example, which is more positive). The annotations can then be converted into a ranking of items by the property of interest, and one can even obtain real-valued scores indicating the degree to which an item is associated with the property of interest.
The paired comparison method does not suffer from the problems discussed above for the rating scale, but it requires a large number of annotations---order $N^2$, where $N$ is the number of items to be annotated.\\[-10pt]

\noindent {\it Best--worst scaling (BWS)} \cite{Louviere_1991} is a form of comparative annotation, like paired comparison, but it requires much fewer annotations.
Annotators are given $n$ items (an $n$-tuple, where $n > 1$ and commonly $n= 4$).\footnote{At its limit, when $n=2$, best--worst scaling reduces to a {\it paired comparison} \cite{thurstone1927law,david1963method}; However, then a much larger set of tuples need to be annotated (closer to $N^2$). } 
They are asked which item is the {\it best} (highest in terms of the property of interest) and which is the {\it worst} (least in terms of the property of interest).
When working on $4$-tuples, best--worst annotations are particularly efficient because each best and worst annotation will reveal the order of
five of the six item pairs (e.g., for a 4-tuple with items \textit{w, x, y,} and \textit{z}, if \textit{w} is the best, and \textit{z} is the worst, then \textit{w} $>$ \textit{x}, \textit{w} $>$ \textit{y}, \textit{w} $>$ \textit{z}, \textit{x} $>$ \textit{z}, and \textit{y} $>$ \textit{z}).
Real-valued scores of association 
between the items and the property of interest can be determined using simple arithmetic on the number of times an item was chosen best and number of times it was chosen worst  
\cite{Orme_2009,flynn2014}.
It has been empirically shown that three annotations each for $2N$ $4$-tuples is sufficient for obtaining reliable scores (where N is the number of items) \cite{Louviere_1991,maxdiff-naacl2016}.
Kiritchenko and Mohammad \shortcite{maxdiff-naacl2016,kiritchenko2017best} showed through empirical experiments on emotion lexicons that BWS produces more reliable and more discriminating scores than those obtained using rating scales. 

Within the NLP community, 
BWS has been used for creating datasets for relational similarity \cite{jurgens-EtAl:2012:STARSEM-SEMEVAL}, word-sense disambiguation \cite{Jurgens2013EmbracingAA}, word--sentiment intensity \cite{maxdiff-naacl2016}, sentence--sentence semantic relatedness \cite{abdalla2023makes}, 
etc.\\[-4pt]

\noindent {\bf \#7. Annotators:} Who is recruited to annotate the data also impacts the lexicon that is generated. \\[-18pt]
\begin{enumerate}[label=\alph*.]
\item \textit{Experts or Crowd:} If a task has clear correct and wrong answers and knowing the answers requires some training/qualifications, then one can employ domain experts to annotate the data. However, emotion annotations largely do not fall in this category. People are the best judges of their emotions and how they use words to communicate them. If the goal is to determine how people use language  or we want to know how people perceive words, phrases, and sentences then we might want to employ a large number of annotators (crowdsourcing). 
Note that this is also a scenario where there can be more than one appropriate answer.\\[-20pt]
\item \textit{Diversity:} Emotion lexicons are a function of their annotators. Consider who all should be represented in the annotator pool, and actively recruit people from under-represented groups. Seek appropriate demographic information (respectfully and ethically).  Document annotator demographics at an aggregate level. \\[-20pt]
\item {\it Informed Consent, Privacy, and Potential for Harms:} Provide a clear and easy-to-understand description of what the task will involve, potential risks, and what information will be collected, before obtaining consent from the annotators. Note that if the terms included for annotation or the chosen dimension of annotation is particularly negative, then there may be significant risk of adversely impacting the annotator's mental health. In such cases, suitable avenues for recourse must be provided.\\[-20pt]
\item \textit{Remuneration:} Determine fair compensation for the task. Inform the annotators of the pay and the time commitment expected.\\[-20pt]
\item \textit{Miscellaneous:}  
There are several other ethical considerations also involved with such work such as: worker invisibility, lack of learning trajectory, humans-as-a-service paradigm, worker well-being, and worker rights
\cite{dolmaya2011ethics,fort-etal-2011-last,standing2018ethical,irani2013turkopticon}.\\[-18pt]
\item \textit{Ethics Approval:} Obtain approval of the project and annotation plan from your institution's research ethics board before conducting the annotation. The ethics boards are also a great source of feedback for improving the ethical standards of the annotation process.
If unsure whether some work requires ethics approval, reach out to the ethics board. Many institutions provide expedited review in cases of low risk.\\[-18pt]
\end{enumerate}
\noindent Document 
these considerations 
so that the users 
can judge suitability of the lexicon for their work.\\[-4pt]

\noindent {\bf \#8. Quality Control:} Good quality control strategies can make a large difference for any scenario of annotations, but are especially important when the annotations are done via crowdsourcing. Quality control strategies can be of three kinds:\\[4pt]
\hspace*{3mm} \textit{Type 1:} applied before data annotation begins\\[2pt] 
\hspace*{3mm}  \textit{Type 2:} applied during data annotation, and\\[2pt]
\hspace*{3mm}  \textit{Type 3:} applied after data annotation.\\[4pt]
\noindent It is recommended to apply measures of all three kinds. Examples of Type 1 include:  careful questionnaire design and setting up training or qualification annotations to screen annotators.

A particularly powerful example of a Type 2 measure is to intersperse the instances with small number  of hidden gold instances ($\sim$5\%) --- instances for which the appropriate label(s) are pre-determined (by, say, the authors). 
If a crowd worker responds with an answer not already marked as appropriate, then they are immediately notified, the annotation is discarded. If an annotator's accuracy on the gold questions falls below a pre-chosen threshold (say, 80\%), then they are refused further annotation, 
and all of their annotations are discarded. 
This way the gold instances serve as a mechanism to avoid malicious annotations, 
as well as a way to further train the annotators. This also avoids scenarios where an annotator provides responses to a large number of questions, only to later learn that they misinterpreted something, rendering all of their annotations useless.
The use of gold questions was popularized by the crowdsourcing platform CrowdFlower (now, Figure8). 

Examples of Type 3 quality control measures include: removal of responses from people who answer questions too quickly, or whose responses are more than two standard deviations away from the responses of others. There also exist approaches that identify which annotators to trust using machine learning algorithms \cite{raykar2012eliminating,hovy-etal-2013-learning}. \\[-3pt]

\noindent \textbf{C. ENTRIES IN THE LEXICON}\\[-8pt]

\noindent {\bf \#9. Annotation Aggregation:} Each instance in a lexicon (usually a word) is often annotated by a number of annotators. Standard practice in aggregating the responses from multiple annotators is to take the most frequent response. However, it should be noted that sometimes other responses are also appropriate. Further, different socio-cultural groups can perceive language differently, and taking the majority vote can have the effect of only considering the perceptions of the majority group. When these views are crystallized in the form of a lexicon, it can lead to the false perception that the norms so captured are ``standard" or ``correct", whereas other associations are ``non-standard" or ``incorrect". Thus, it is worth explicitly disavowing that view and stating that the lexicon simply captures the perceptions of the majority group among the annotators. 
Thus, it is recommended to also make available disaggregated annotations (annotations in their raw form -- without aggregation). 
Note that it is also problematic to consider 
all annotator responses as valid because sometimes annotators make mistakes, and some may have inappropriate biases (see \#15).\\[-10pt]

\noindent \textbf{\#10. Relative (not Absolute):} The absolute values of the association scores themselves usually have no meaning. The scores help order the words relative to each other. For example, a term with a high valence score is associated with more positiveness than a term with with a lower score. \\[-10pt] 

\noindent \textbf{\#11. Coverage:} Some lexicons have a few hundred terms, and some have tens of thousands of terms. However, even the largest lexicons do not include all the terms in a language. 
Mostly, they include entries for the canonical forms (lemmas), but some also include morphological variants. 
The high-coverage lexicons, such as the NRC Emotion Lexicon, 
have tens of thousands of terms. 
However, when using the lexicons in specialized domains, one may find that a number of common terms in the domain are not listed in the lexicons. 

\noindent \textbf{\#12. Not Immutable:}
The associations do not indicate an inherent unchangeable attribute. 
Emotion associations can change with time, but these lexicon entries are largely fixed. They pertain to the time they are created or the time associated with the corpus from which they are created. \\[-10pt]

\noindent \textbf{\#13. Perceptions (not ``truth''):}
Emotion lexicons largely capture how speakers of a language perceive the emotion associations of words. 
As mentioned in the previous bullet, this can change with time. Further, it can also be different for different people. Mohammad and Turney \shortcite{MohammadT13} found that when the annotators are asked to judge emotion associations in terms of `how speakers of a language perceive the word', the results have lower variance than when asked `the emotions evoked in the annotator'. Consider your objective when deciding which of the two framings (or some other) is more appropriate for your use case. \\[-10pt]

\noindent \textbf{\#14. Socio-Cultural Biases:}
Since the emotion lexicons have been created by people (directly through crowdsourcing or indirectly through the texts written by people) they capture various human biases.
These biases may be systematically different for different socio-cultural groups. Document who produced the data (people from which countries, what is the gender distribution, age distribution, etc.) in the paper describing the dataset or in the associated datasheet. An advantage of crowdsourcing is that the annotations are from a wider pool of annotators; however, crowd annotators are systematically different from, and not representative of, the general population.\\[-10pt] 

\noindent \textbf{\#15. Inappropriate Biases:} Some of the human biases that have percolated into the lexicons may be rather inappropriate. For example, entries with low valence scores for certain demographic groups or social categories. 
Studying such biases in the lexicon can be useful to show and address some of the historical inequities that have plagued humankind. Nonetheless, when these lexicons are used in specific tasks, care must be taken to 
remove such entries from the lexicons where necessary.\\[-10pt]

\noindent \textbf{\#16. Errors:}
Even though the researchers take several measures to ensure high-quality and reliable data annotation (e.g.,
multiple annotators, clear and concise questionnaires, framing tasks as comparative annotations,
interspersed check questions, etc.), human-error can never be fully eliminated in large-scale annotations.
Expect a small number of clearly wrong entries.
Automatically generated lexicons also can have erroneous entries. They are often built on the assumption that the tendency of a word to co-occur with emotion-associated seed terms is proportional to its association with that emotion. However, in any corpus, there will always be some amount of chance high co-occurrences that are not accurate reflections of the true associations.\\[-10pt]


\noindent \textbf{\#17. Mechanism to Report and Fix Errors:} Provide a mechanism for users to report issues and errors. Fix errors and where appropriate issue warnings for how some types of entries can be mis-interpreted or misused. Periodically assess whether certain types of entries need to be proactively checked. For example, there has been growing recognition that emotion associations associated with identity groups are particularly sensitive, affected by historical bias, and so one must be careful in how they interpret the associations captured in lexicons.\\

\noindent \textbf{D. APPLYING THE LEXICON}\\[-8pt]



\noindent {\bf \#18. Examining the Fit of the Lexicon:} Manually examine the emotion associations of the most frequent terms in your data. Remove entries from the lexicon that are not suitable (due to mismatch of sense, inappropriate human bias, etc.).\\[-10pt]

\noindent {\bf \#19. Rescaling the Lexicon for One's Task:} Depending on your specific use case, you may choose to re-scale the scores from 0 to 1, -1 to 1, 1 to 10, etc. Note that if using the lexicon entries as features in machine learning experiments, the scale (0 to 1 or -1 to 1) can make a difference---e.g. if the score is used as a weight for features.\\[-10pt]

\noindent {\bf \#20 Metrics and Features Drawn from the Lexicon:} For text analysis, one can calculate various metrics such as the percentage of emotion words (when the lexicons provides a list of words associated with a category) or average emotion intensity (for real-valued associations).
When determining the scores, a further choice is how to handle words that are not in the lexicon.
Two common approaches include: 1. Treat words that are not in the lexicon as neutral; 
2. Ignore these words in the calculation of the scores. The latter approach does not make assumptions of neutrality, and is not impacted by the number of such
out of lexicon words in a piece of text.
See \newcite{TMemotionarcs} for a systematic analysis of the impact of various lexicon features on the quality of emotion arcs generated with them.\\[-10pt]

\noindent {\bf \#21. Creating Subsets of the Lexicon:} Sometimes it is better to use a subset of the emotion lexicon, rather than the whole lexicon.\\[3pt] 
\noindent \textit{Removing Neutral Words:} 
One can use the whole lexicon to calculate 
metrics such as average valence of the words in a text;
however, one can also choose to disregard terms with close to 0 valence scores. 
when calculating the same metric.
Removal of such neutral terms from the analysis will show greater variations in the average scores when comparing across different sets of data of interest  
or across time. For example, when looking at the average tweet happiness over time of day, using full or neutral-removed lexicon is expected to get roughly similar curves, but the neutral-removed lexicon will show a greater amplitude (divergence of scores from the peaks to troughs).
\cite{dodds2011temporal} describes this as turning up the magnifier knob in a microscope.
\sm{Note, however, that just having larger score differences between the target and control does not mean that the emotion word usage is substantially different or significant; and conversely, just because the score difference for a metric is small in value does not mean that the differences in emotion word usages are not substantial. 
(More on this in \#22).}\\[3pt]
\noindent \textit{Removing Low-Association Words:} Use of low-association terms from a lexicon may not be beneficial for some downstream applications. These entries may also include a greater percentage of annotation errors.
See \newcite{TMemotionarcs} for experiments on multiple datasets and multiple emotion dimensions that examine usefulness of removing low-association terms from a lexicon when generating emotion arcs.\\[3pt]
\noindent \textit{Removing Highly Polysemous and Certain Domain Words:} For some applications, it is beneficial to discard highly ambiguous words. 
Entries for highly ambiguous words are more likely to include emotion associations for a sense that is not common in one's data.
As stated in \#3, it is also recommended to remove entries not appropriate for the target domain; e.g., 
the word {\it harry} has a negative meaning, but it should not be used when analyzing text where 
a person has the name \textit{Harry}.\\[-10pt]


\noindent {\bf \#22. Inferences:} When drawing inferences from texts using counts of emotion words:\\[-19pt]
\begin{enumerate}[label=\alph*.]
\item It is more appropriate to make claims about emotion word usage rather than emotions of the speakers. For example, {\it `the use of anger words grew by 20\%'} rather than {\it `anger grew by 20\%'}. 
A marked increase in anger words is likely an indication that anger increased, but there is no evidence that anger increased by 20\%.
\sm{Further, it is important to understand the emotion metrics and to interpret them accordingly.
For example, many off-the-shelf tools provide a ``sentiment score" for the input textual instances, without  providing
adequate details about what this score means. As discussed in \#21, the scores themselves can have large or small values, and just knowing that
the score difference between a target and control is large (or small) is not enough to draw meaningful inference. On the other hand,
grounded metrics that tie the score to attributes such as percentage of positive words tend to be less open to misinterpretation.}
\\[-22pt]
\item Comparative analysis is your friend. Often, emotion word counts on their own are not useful. 
For example, {\it `the use of anger words grew by 20\% when compared to [data from last year, data from a different person, etc.]'} is more useful than saying {\it `on average, 5 anger words were used in every 100 words'}.\\[-22pt]
\item Lexicon features (or any other automatically drawn features) are \textit{not} well suited to draw meaningful emotional inferences from individual utterances. Human language and behaviour are highly variable and complex. However, with careful design, they can be
useful to draw inferences about broad trends at an aggregate level \cite{TMemotionarcs}. \\[-22pt]
\item Inferences drawn from large amounts of text are more reliable than those drawn from small amounts of text. \newcite{TMemotionarcs} show that this is the single most important feature in determining the fidelity of the predicted emotion trends with the true emotion trends,
among a host of features they explored. For many emotion dimensions and dataset domains, it is advisable to determine aggregate emotion scores using at least 100 instances. For example, if there are at least 100 tweets per day about a product of interest, the average valence scores of all the words in the tweets every day is expected to produce a fairly accurate valence arc (x-axis is day, y-axis is average valence score for the corresponding day). 

\end{enumerate}




\section{Limitations}
This paper does not present a new NLP model or dataset. Thus, there are no corresponding limitations to discuss. 
However, the paper itself can be viewed as a document discussing limitations of existing approaches to do sentiment and emotion analysis using emotion lexica. The 22 best practises presented in the paper 
discuss approaches to engage with and counter these limitations.

While this document was a result of engaging a larger community through blog posts, talks, and discussions, we had relatively low access to developers of commercial sentiment analysis systems.
Thus the list presented here may have missed some important considerations.
We encourage readers and impacted stakeholders to challenge the assumptions latent in the document, and identify new ethical considerations not included here or not gaining adequate attention in the research community.

\section{Concluding Remarks}
\noindent Emotion lexicons are simple yet powerful tools to analyze text. However, use of the lexicons (even for tasks that it is suited for) can lead to inappropriate bias. Applying a lexicon to any new data should only be done after first investigating its suitability, and requires careful analysis to minimize unintentional harm. 
In this paper, we presented 22 best practises that include considerations that can help mitigate such unwanted outcomes, as well as strategies to make the best use of emotion lexicons towards drawing meaningful and accurate inferences.
The best practises are organized as per a lexicon's life cycle:
A. Lexicon Design, B. Annotation, C. Entries in the Lexicon, and D. Applying the Lexicon. 
We also provide pointers to relevant literature to explore the best practises in more detail.
It should be noted that these practises are not meant to be the final word, but rather jumping off points for further thought, discussion, and additional measures towards the responsible use of emotion lexicons.

\section*{Acknowledgments}
 Many thanks to Emiel van Miltenburg, Annika Schoene, Mallory Feldman, Tara Small, Roman Klinger,  and Peter Turney for thoughtful comments and discussions.


\bibliography{custom}

\bibliographystyle{acl_natbib}




\end{document}